\documentclass[lettersize,journal]{IEEEtran}
\usepackage{amsmath,amsfonts}
\usepackage{algorithmic}
\usepackage{algorithm}
\usepackage{array}
\usepackage[caption=false,font=normalsize,labelfont=sf,textfont=sf]{subfig}
\usepackage{textcomp}
\usepackage{stfloats}
\usepackage{url}
\usepackage{verbatim}
\usepackage{graphicx}
\usepackage{cite}

\usepackage{multirow}
\usepackage{bbding}  
\usepackage{booktabs} 
\usepackage{colortbl} 
\usepackage{makecell}
\usepackage{diagbox}
\usepackage{xcolor}
\usepackage{hyperref}
\usepackage{tikz}
\usepackage[switch]{lineno}

\hypersetup{
colorlinks=true,
linkcolor=black,
citecolor=black,
}

\begin{document}
\title{Causal Clothes-Invariant Feature Learning for Cloth-Changing Person Re-ID}

\author{Xulin Li, Yan Lu, Bin Liu, 
Jiaze Li,
Yating Liu, Qi Chu, Mang Ye, Wanli Ouyang,
Nenghai Yu
\thanks{This work was supported by the National Natural Science Foundation of China under Grant 62272430. Xulin Li and Yan Lu contributed equally to this work. Corresponding author: Bin Liu.}
\thanks{
Xulin Li, Bin Liu, Jiaze Li, Qi Chu, and Nenghai Yu are with the School of Cyber Science and Technology, University of Science and Technology of China, Hefei 230026, China, and also with Anhui Province Key Laboratory of Digital Security, Hefei 230026, China (e-mail: lxlkw@mail.ustc.edu.cn; flowice@ustc.edu.cn; jz\_li@mail.ustc.edu.cn; qchu@ustc.edu.cn; ynh@ustc.edu.cn).
Yan Lu and Wanli Ouyang are with Shanghai Artificial Intelligence Laboratory, Shanghai 200232, China (e-mail: luyan@pjlab.org.cn; wlouyang@ie.cuhk.edu.hk).
Yating Liu is with the School of Data Science, University of Science and Technology of China, Hefei 230026, China (e-mail: liuyat@mail.ustc.edu.cn).
Mang Ye is with Wuhan University, Wuhan 430072, China (e-mail: yemang@whu.edu.cn).
}
\thanks{Accepted version. IEEE Transactions on Circuits and Systems for Video Technology, doi: 10.1109/TCSVT.2026.3700883.}
}

\IEEEpubid{%
\begin{minipage}{0.96\textwidth}
\centering\scriptsize
Copyright \copyright\ 2026 IEEE. Personal use of this material is permitted.
However, permission to use this material for any other purposes must be obtained from the IEEE by sending an email to pubs-permissions@ieee.org.
\end{minipage}}

\maketitle

\begin{abstract}

In cloth-changing person re-identification (CC-ReID), it is critical to learn clothes-invariant feature, which can provide discriminative ID features that remain robust against clothing changes.
However, a spurious correlation currently limits existing ReID methods from effectively extracting these clothing-invariant features.
This spurious correlation arises from clothing ownership: clothing is rarely shared across different identities, so models tend to memorize clothing cues for identity recognition, and this strategy generalizes poorly to unseen clothing.
In this paper, we propose Causal Clothes-Invariant Learning (CCIL), which explicitly shifts CC-ReID from likelihood learning $P(Y|X)$ to causal intervention learning $P(Y|do(X))$ to block the clothing shortcut.
CCIL realizes this intervention through three modules: a Confounder Dictionary, an Intervention Module, and Disentangle Regularization.
The causality-based modeling makes the entire model naturally clothes-invariant, effectively preventing the capture of spurious correlations in feature learning.
Extensive experiments validate the effectiveness of CCIL. On PRCC and DeepChange datasets, CCIL achieves Rank-1 accuracies of 66.4\% and 59.2\%, outperforming state-of-the-art methods by 1.4 and 4.1 percentage points, respectively.
\end{abstract}

 
\section{Introduction}
\label{sec:intro}
\IEEEPARstart{P}{erson} re-identification (ReID) aims to retrieve specific pedestrians across different cameras, which is widely used in city surveillance, intelligent security, and related areas.
It is challenging due to the existence of frequent occlusion\cite{hou2021feature}, background interference~\cite{tian2018eliminating}, and illumination variations\cite{zhang2022illumination}.
Significant progress~\cite{sun2018beyond,ye2021deep,leng2019survey} has been observed in standard person ReID, which assumes persons do not change their clothes throughout the entire retrieval process.
To meet the long-term retrieval requirements of real-world scenarios, cloth-changing ReID (CC-ReID)~\cite{qian2020long,yang2019person} takes into consideration the extra challenge of clothing variations.

\begin{figure}[t]
    \centering
    \includegraphics[width=0.9\linewidth]{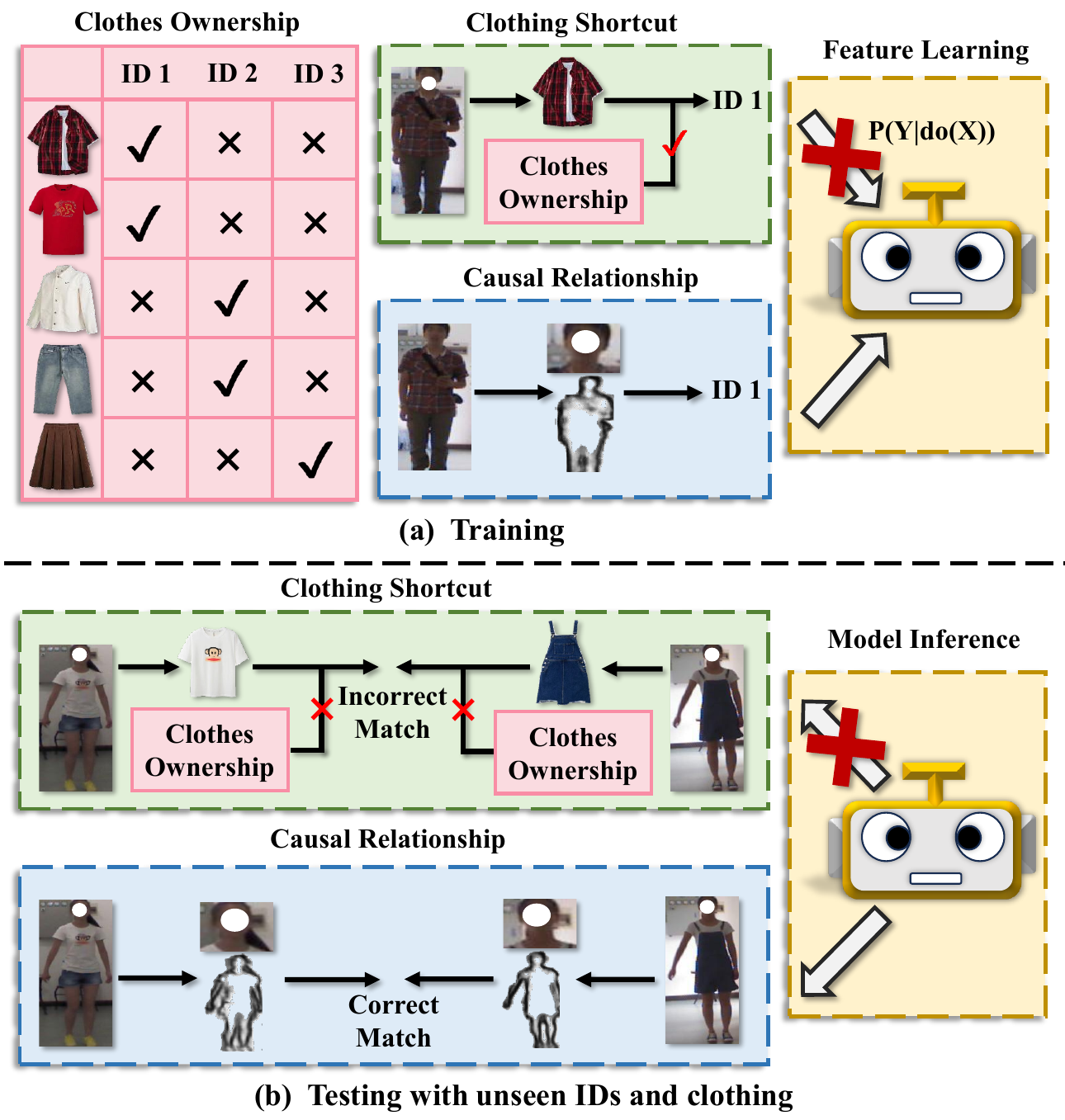}
    \caption{
    The spurious correlation between clothing and IDs can lead to a clothing shortcut of identifying persons based on their clothing.
    This shortcut is not a robust recognition process as it fails when encountering IDs and clothing not in the training set.
    Our approach uses the causal intervention $P(Y|do(X))$ to enable the deep model to solely learn the causal relationship based on discriminative ID clues, thus demonstrating good generalizability in testing scenarios.
    }
    \label{fig:intro}
    \vspace{-3mm}
\end{figure}

\IEEEpubidadjcol

Whatever specific ReID tasks are, invariant feature learning is critical, which guides the model to extract invariant cues and adapt to complex scenarios.
In the CC-ReID task, clothes-invariant feature learning holds great importance as it requires discriminative features that are robust to clothing changes.
Compared with other interfering factors (e.g., scene, illumination, camera, and viewpoint variations), clothing changes are particularly challenging for invariant feature learning. This is because clothing and human IDs exhibit strong spurious correlations, while correlations between other interfering factors and identity are typically much weaker.

As shown in Figure~\ref{fig:intro}, the spurious correlation is caused by a fact that each outfit is only worn by its owner and hardly shared with others.
This kind of clothes ownership is naturally reflected in existing CC-ReID datasets~\cite{yang2019person,qian2020long,xu2023deepchange}, resulting in a clothing shortcut that recognizes people by identifying their wearing clothes.
Existing methods inevitably capture this shortcut because it would help the model achieve training targets more directly.

However, this clothing shortcut is an unreliable relationship.
When we deploy such a model in real applications, facing unknown people wearing unseen clothes before, the model still attempts to utilize the aforementioned clothing shortcut to recognize people, leading to bad re-identification results.

To address this challenge, we proposed a Causal Clothes-Invariant Learning (CCIL), which focuses on learning causal relationships. 
This new causality-based training framework can avoid the clothing shortcut during training, which promotes invariant feature extraction.
The main idea of the proposed CCIL is to model the causal intervention probability~\cite{glymour2016causal,pearl2018book}, denoted as $P(Y|do(X))$, rather than the likelihood probability $P(Y|X)$ commonly modeled by most existing methods, where $Y$ means the human ID and $X$ means the input image. 
As shown in Figure~\ref{fig:intro}, the $P(Y|do(X))$ only models the causal relationships from $X$ to $Y$ and does not include the clothing shortcut, while $P(Y|X)$ models all relationships between $X$ and $Y$.
The causal relationship reflects the stable association of how to infer human identity from a given image by using discriminative ID clues, making the models applicable to various complex scenarios.
So the CCIL which models the $P(Y|do(X))$ could capture more stable causal patterns and neglect the spurious correlation, leading to better clothes-invariant representation for CC-ReID. 

To make the CCIL model the $P(Y|do(X))$ better within a CC-ReID framework, we design our CCIL by following three novel modules.
Firstly, we learn the representation distribution of clothing in the dataset and store them in a Confounder Dictionary to support subsequent causal intervention implementation.
Secondly, the Intervention Module employs a novel and effective approach to model the backdoor adjustment formula in causal theory, leading to the derivation $P(Y|do(x))$.
Lastly, the Disentangle Regularization improves the modeling of clothing in the Confounder Dictionary, thereby further enhancing the effectiveness of the Intervention Module.
The overall feature learning process is combined within the causal intervention, resulting in clothes-invariant feature extraction.

Our main contributions are summarized as follows:  

\noindent $\bullet$ We analyze the barrier of clothes-invariant feature learning in CC-ReID, which is the clothing shortcut in the training set, and propose a novel framework dubbed Causal Clothes-Invariant Learning (CCIL) via the causal intervention view.
Our CCIL first provides a new causal solution based on backdoor adjustment for the CC-ReID task, which focuses on eliminating the clothing shortcut.

\noindent $\bullet$ To achieve the CCIL, we present three causal modules to implement confounder modeling, causal intervention, and disentangled regularization, to jointly model $P(Y|do(X))$.
With our CCIL, deep models are guided to overlook the clothing shortcut and instead capture more discriminative ID clues.

\noindent $\bullet$ Extensive experiments on multiple CC-ReID datasets validate the effectiveness and superiority of our method against the state-of-the-art causal and non-causal ReID methods in mitigating the clothing shortcut.

\section{Related Work}
\label{sec:relate}
{\bf Cloth-Changing Person ReID.}
Person re-identification (ReID) is the task of retrieving persons of interest across non-overlapping cameras.
However, standard ReID methods~\cite{zheng2015scalable,wei2018person} face limitations in long-term scenarios where individuals change their clothes over time. 
Consequently, there has been a growing interest in the field of cloth-changing person re-identification (CC-ReID)~\cite{yang2019person,qian2020long,xu2023deepchange,wang2020benchmark,huang2019beyond,shu2021large,pang2025identity} in recent years.

Most CC-ReID methods primarily use extra clothing-agnostic modalities data to guide model training.
Yang et al.~\cite{yang2019person} used pure contour sketches for discriminative feature learning.
Chen et al.~\cite{chen2021learning} directly extracted a texture-insensitive 3D shape embedding from a 2D image by adding 3D body reconstruction as an auxiliary task.
Hong et al.~\cite{hong2021fine} used 2D silhouettes, Qian et al.~\cite{qian2020long} introduced key points, Jin et al.~\cite{jin2022cloth} and Lu et al.~\cite{lu2024flag} utilized gait, Cui et al.~\cite{cui2023dcr} and Xiong et al.~\cite{xiong2025hprnet} used human parsing, Li et al.~\cite{li2022cocas+} introduced clothes templates to assist robust features learning to clothes change.
Other methods only use the original RGB image to solve CC-ReID.
Huang et al.~\cite{huang2019beyond} and Shu et al.~\cite{shu2021large} improved CC-ReID from the network architecture and ranking loss perspectives, respectively.
Gu et al.~\cite{gu2022clothes} introduced an adversarial loss to decouple clothes-irrelevant features from the RGB modality.
Han et al.~\cite{han2023clothing} proposed clothing-change augmentation methods to address the limitation of the insufficient number and variation of clothing in training data.

These methods are essentially likelihood-based and optimize $P(Y|X)$, which inevitably captures the clothing shortcut. Even methods using clothing-agnostic auxiliary modalities do not discard RGB images, and the impact of spurious correlation is diluted but not eliminated.
In contrast, our CCIL models $P(Y|do(X))$ via backdoor adjustment to directly cut off the clothing shortcut through causal intervention.

{\bf Causal Inference in Person ReID.}
Thanks to the powerful ability of causal inference to remove bias and pursue causal effects, existing research has explored the incorporation of causal inference in person ReID.
Rao et al.~\cite{rao2021counterfactual} proposed a method for counterfactual attention learning to enhance the attention module.
Li et al.~\cite{li2022counterfactual} utilized the total indirect effect (TIE) to emphasize the significance of graph topology in cross-modality ReID task.
These methods focus on enhancing specific ReID modules, such as attention and graph modules, while our method is model-agnostic.
Zhang et al.~\cite{zhang2022learning} achieved domain-invariant representation learning through approximated causal interventions.
Domain-invariant learning cannot be applied to achieve clothes-invariant learning due to their involving different confounders, domains, and clothing.
Consequently, we adopted different techniques for modeling the confounder and implementing the causal intervention.

In CC-ReID, Yang et al.~\cite{yang2023good} utilized the total direct effect (TDE) to highlight the direct effect (image $\rightarrow$ identity).
In contrast, we employ the backdoor adjustment method to guide the model in capturing purer clothing-invariant features and abandoning clothing-related shortcuts (image $\leftarrow$ clothing $\rightarrow$ identity) as much as possible, which results in more robust CC-ReID results.

{\bf Causal Inference in Other Tasks.} 
Recently, improving deep learning through causal inference~\cite{pearl2009causal,yao2021survey} has received increasing attention.
It has been applied to various fields, including categorization~\cite{tang2020long}, visual question answering~\cite{niu2021counterfactual}, semantic segmentation~\cite{zhang2020causal}, object detection~\cite{wang2020visual}, large language models~\cite{kiciman2023causal}.
The backdoor adjustment~\cite{pearl2018book,glymour2016causal} is a causal inference theory to achieve the causal intervention $P(Y|do(X))$.
Some deep learning methods~\cite{wang2020visual,zhang2022multiple,zhang2020causal,li2021interventional,lin2022causal,huang2023causalainer} model confounders using the mean of features, and subsequently simulate backdoor adjustment through cross-attention to mitigate the effects of confounders from their corresponding tasks.
We employed the backdoor adjustment algorithm for the first time in CC-ReID with a novel approach for confounder modeling and causal intervention implementation to perform invariant feature learning.
 
\section{Causal Analysis}
\label{sec:causal}
In this section, we introduce causal theory to analyze the negative effect caused by the clothing shortcut in the CC-ReID task.
We demonstrated that causal intervention can be achieved by modeling $P(Y|do(X))$, which theoretically tackles the clothing shortcut problem, thereby facilitating the extraction of clothes-invariant feature.

\subsection{Structural Causal Model For CC-ReID}

\begin{figure}[t]
    \centering
    \includegraphics[width=1.0\linewidth]{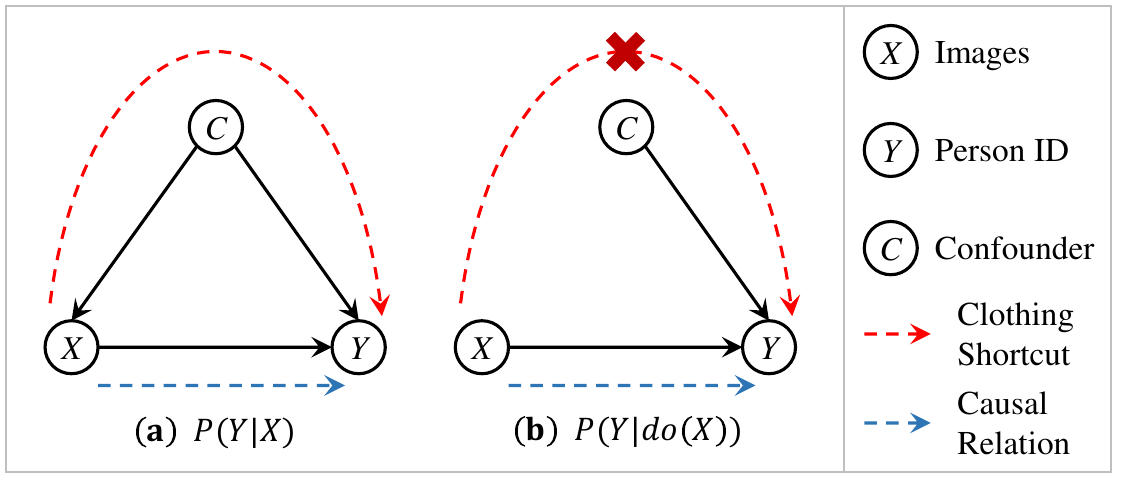}
    \caption{
    Causal graphs for CC-ReID.
    (a) The clothing shortcut $X\leftarrow C\rightarrow Y$ captured by the likelihood-based methods that directly utilize $P(Y|X)$ to model the relationships from $X$ to $Y$.
    (b) Our method models the relationships by the causal intervention probability $P(Y|do(X))$, which only captures the causal relation $X\rightarrow Y$ and removes the clothing shortcut.
    }
    \label{fig:causal}
    \vspace{-4mm}
\end{figure}

We analyze the causality in the CC-ReID task by using a Structural Causal Model (SCM)~\cite{glymour2016causal,pearl2018book}.
The SCM is built to depict the causal relationships among the variables `images' $X$, `human ID' $Y$, and confounder `clothes' $C$.
As shown in Figure~\ref{fig:causal} (a), the solid arrows denote the causal relationships: cause $\rightarrow$ effect.
$X\rightarrow Y$ denotes the labeling process from the given image, as an image is labeled for its content.
An ideal CC-ReID model only identifies $X\rightarrow Y$, which is unbiased.
$C\rightarrow X$ indicates that different clothes result in diverse image contents. 
$C\rightarrow Y$ implies that the identity can be inferred from the clothes, depicting the presence of the spurious correlation caused by data collection.
Other interfering factors (e.g., scene, camera, and viewpoint) usually cannot be used to infer identity, and thus do not open backdoor paths. Therefore, they are not considered confounders under the causal inference framework.
In summary, there are two relationships from $X$ to $Y$:
a clothing shortcut $X\leftarrow C\rightarrow Y$ (also known as the backdoor path) and a causal relation $X\rightarrow Y$. 

\subsection{Analysis of Existing Likelihood-Based Methods}
The likelihood-based method is influenced by the clothes-identity spurious correlation, thus capturing a clothing shortcut for identifying persons through their clothing.
As shown in Figure~\ref{fig:causal} (a) likelihood-based methods directly learn the probability $P(Y|X)$ to model the correlation relationship between the $X$ and $Y$ by deep models.
However, the clothing shortcut and the causal relationship are entangled together, directly learning $P(Y|X)$ will capture the clothing shortcut unavoidably. 
Besides, because of such kind of abstract modeling approach, the corresponding patterns of the clothing shortcut will be just represented in an implicit and unexplainable way.
Therefore, once the clothing shortcut $X\leftarrow C\rightarrow Y$ is captured by the model, removing the corresponding patterns from the model without affecting the useful knowledge related to the causal relationships proves to be quite challenging.

\subsection{Causal Intervention Helps Invariant Feature Learning}
In theory, causal intervention can cut off the clothing shortcut and learn the relationship unaffected by clothing, which is consistent with the clothes-invariant objective.
The intervention operation can be defined at the \textit{Do}-operation~\cite{pearl2000models} $do(\cdot)$, denoting the causal relationship between $X$ and $Y$ as $P(Y|do(X))$.
As shown in Figure~\ref{fig:causal} (b), the intervention disrupts the potential correlation between the intervening variable $X$ and its cause $C$, depicted as the removed arrow $X \leftarrow C$.
In this case, there is only the causal relation $X\rightarrow Y$ between the $X$ and $Y$.
However, achieving intervention is not straightforward.
Fortunately, the `backdoor adjustment' theory~\cite{pearl2018book,glymour2016causal} allows us to compute the intervention probability\footnotemark[1]:
\begin{equation}
\label{eq:int_prob}
\begin{aligned}
P(Y|do(X))=\sum\limits_c P(Y|X, c)\cdot P(c).
\end{aligned}
\end{equation}
\footnotetext[1]{The detailed proof is provided in the Supplementary Material.}
In comparison to the original likelihood $P(Y|X)$ in the Bayesian framework:
\begin{equation}
\label{eq:lik_prob}
\begin{aligned}
P(Y|X)=\sum\limits_c P(Y|X, c)\cdot P(c|X),
\end{aligned}
\end{equation}
it is evident that the intervention modified $P(C=c|X)$ into a prior probability $P(C=c)$, which is equal to making clothes $C$ statistically independent of images $X$. 
In this case, the backdoor path $X\leftarrow C\rightarrow Y$ is removed.
As a result, the clothing shortcut that identifies persons based on their attire is not applicable and the model will learn undisturbed to capture clothes-invariant features.
 
\begin{figure*}[htbp]
    \centering
    \includegraphics[width=0.90\linewidth]{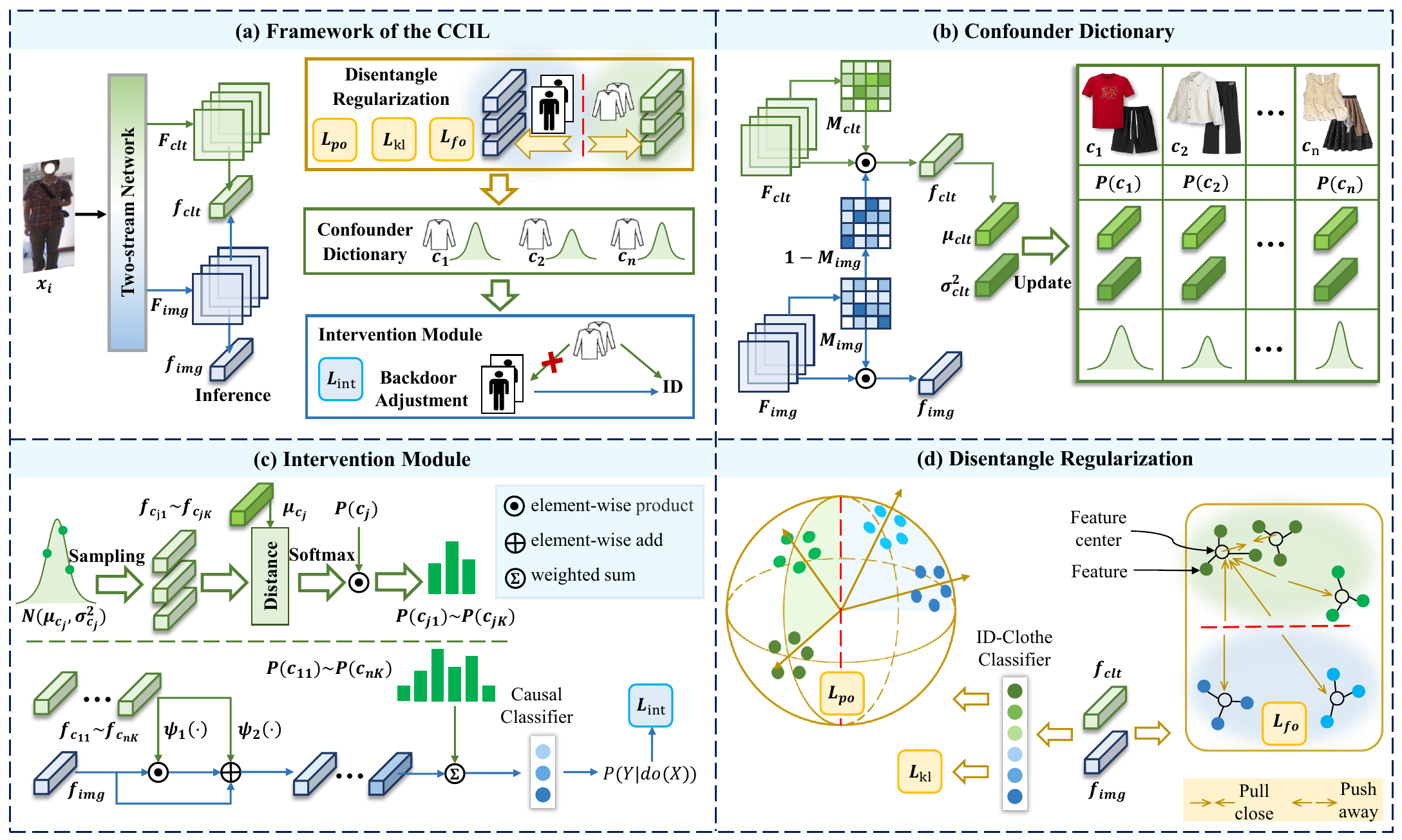}
    \caption{
    (a) The framework of the proposed Causal Clothes-Invariant Learning (CCIL). 
    Three causal modules are complementary to each other and train the model under the causal intervention framework, achieving better clothes-invariant features.
    (b) The Confounder Dictionary models clothing to support the implementation of the subsequent causal intervention.
    (c) The Intervention Module collaborates with the established Confounder Dictionary to derive the $P(Y|do(x))$.
    (d) The Disentangle Regularization improves the modeling of clothing in the Confounder Dictionary, thereby further enhancing the effectiveness of the Intervention Module.
    }
    \label{fig:method}
  \vspace{-3mm}
\end{figure*}

\section{Method}
\label{sec:method}

{\bf Problem Formulation.}
For a CC-ReID dataset $\mathcal{G}=\{(x_i,y_i,c_i)\}_{i=1}^{N_{img}}$, the i-$th$ data sample in $\mathcal{G}$ can be denoted as a triplet $(x_i,y_i,c_i)$, where $x_i$, $y_i$, $c_i$ denotes the image, identity label and the clothing label, respectively.
The CC-ReID task focuses on extracting clothes-invariant and discriminative image features for inference.

{\bf Method Overview.}
The main pipeline of our Causal Clothes-Invariant Learning (CCIL) is shown in Figure~\ref{fig:method}.

\noindent $\bullet$ During training, a given image would be fed into a two-stream network to obtain image features and clothing features.
The clothing features are utilized to establish a Confounder Dictionary iteratively (Sec.~\ref{sec:confounder}).

\noindent $\bullet$ With this, the Confounder Dictionary stores richer information about each cloth, which then interacts with the image features to model causal intervention probability $P(Y|do(X))$  (Sec.~\ref{sec:intervention}).

\noindent $\bullet$ To further make the modeled intervention $P(Y|do(X))$ more accurate, three losses are proposed to achieve clothes and identity disentangling, which improves the modeling of clothing in the Confounder Dictionary and enhances the effectiveness of the Intervention Module (Sec.~\ref{sec:disentangle}).

The aforementioned three modules collectively implement the causal intervention during training, leading to discriminative clothes-invariant feature learning.

\subsection{Confounder Dictionary}
\label{sec:confounder}

We construct a Confounder Dictionary $\mathcal{D}$ to represent all $N_{clt}$ pieces of clothing in the training set, supporting the implementation of causal interventions in subsequent Equation~\ref{eq:pydox}.
As shown in Figure~\ref{fig:method} (b), the Confounder Dictionary models each clothing item $c_i$ as a multivariate Gaussian distribution $\mathcal{N}(\mu_{c_i},\sigma^2_{c_i})$, where $\mu_{c_i}\in\mathbb{R}^{d}$ and $\sigma^2_{c_i}\in\mathbb{R}^{d}$ represent the mean and variance, respectively.
Modeling the distribution of each clothing item can significantly reduce storage overhead compared to storing the clothing features of all images, and it also provides more information than simply storing the mean clothing features.
Moreover, the Confounder Dictionary also stores the probability of each clothing, denoted as $P(c_i)$, which can be calculated in the training set. 
The Confounder Dictionary can be defined as follows:
\begin{equation} 
\begin{aligned}
\mathcal{D}=\{(\mathcal{N}(\mu_{c_i},\sigma^2_{c_i}),P(c_i)\}_{i=1}^{N_{clt}}.
\end{aligned} 
\end{equation}
We calculate the mean $\mu_{clt}$ and variance $\sigma^2_{clt}$ of clothing features $f_{clt}$ within the training
batch.
\begin{equation}  
\begin{aligned}  
\label{eq:mu}        
\mu_{clt} &= \frac{1}{N(c=c_i)}\sum_{c=c_i} f_{clt},\\ 
\sigma_{clt}^2 &= \frac{1}{N(c=c_i)}\sum_{c=c_i} (f_{clt} - \mu_{clt})^2,
\end{aligned}  
\end{equation}
where $N(c=c_i)$ represents the number of samples with clothing label $c_i$ in each training batch.
Subsequently, based on the clothing labels, we update the relevant $\mu_{c_i}$ and $\sigma^2_{c_i}$ items of the Confounder Dictionary by the exponential moving average (EMA) scheme: 
\begin{equation}
\begin{aligned} 
\label{eq:alpha}       
\mu_{c_i} &= \alpha\cdot \mu_{c_i} + (1-\alpha)\cdot \mu_{clt},\\
\sigma_{c_i}^2 &= \alpha\cdot \sigma_{c_i}^2 + (1-\alpha)\cdot \sigma_{clt}^2,
\end{aligned} 
\end{equation} 
where $\alpha$ denotes the memory coefficient.

In addition, the clothing features $f_{clt}\in \mathbb{R}^d$ used to construct the Confounder Dictionary are extracted through spatial attention, which distinguish them from identity features $f_{img}\in \mathbb{R}^d$: 
\begin{equation}
\label{eq:att}
\begin{aligned}
    &f_{clt} = Pool(F_{clt}\odot M_1(F_{clt})\odot(1-M_2(F_{img}))) \ ,\\
    &f_{img} = Pool(F_{img}\odot M_2(F_{img}) \ ,
\end{aligned}
\end{equation}
where $F_{clt}\in \mathbb{R}^{d\times h\times w}$ and $F_{img}\in \mathbb{R}^{d\times h\times w}$ are the clothing feature map and identity feature map output by the two-stream network. $M_1$ and $M_2$ are implemented by a 2D convolutional layer with an output dimension of 1, followed by sigmoid activation. $\odot$ denotes the element-wise product, $Pool$ denote the pooling operation applied along the spatial axes.
This strategy of multiplying the clothing feature map by the reverse attention mask of the identity feature aids in accurately modeling confounders, ensuring that subsequent causal interventions do not mistakenly eliminate identity-related information.

\subsection{Intervention Module} 
\label{sec:intervention}

To eliminate the interference of the clothing shortcut, an intervention loss $\mathcal{L}_{int}$ is constructed by maximizing the intervention probability:
\begin{equation}
\label{eq:int_op}
\begin{aligned}
\mathcal{L}_{int} = \mathbb{E}\left[-\text{log}\ (P(Y=y_i|do(X=x_i))\right].
\end{aligned}
\end{equation} 
The intervention reflects the causality between variables $X$ and $Y$, uncorrelated with the confounder, which is consistent with the clothes-invariant objective.

To achieve the intervention optimization in Equation~\ref{eq:int_op}, we design an Intervention Module to implement the $P(Y|do(X))$ well within the CC-ReID framework.

As shown in Figure~\ref{fig:method} (c), the Intervention Module takes the image feature $f_{img}$ of image $x_i$ and the Confounder Dictionary $\mathcal{D}$ as its inputs, and then calculates the intervention probability based on
backdoor adjustment theory~\cite{pearl2018book,glymour2016causal} that we review in Equation~\ref{eq:int_prob}.
With this module, the intervention probability is calculated as follows:
\begin{equation}
\label{eq:pydox}
\begin{aligned}
    P(Y|do(X=x_i)) &=\sum\limits_{c_j} P(Y|X=x_i, c_j)\cdot P(c_j)\\
    &= \sum\limits_{c_j} Cls[g(f_{img},\mathcal{N}(\mu_{c_j},\sigma^2_{c_j}))]\cdot P(c_j),
\end{aligned}
\end{equation}
where $\mathcal{N}(\mu_{c_j},\sigma^2_{c_j})$ and $P(c_j)$ are achieved in the Confounder Dictionary. 
$Cls(\cdot)=Softmax(Linear(\cdot))$ is a classifier defined as a linear layer followed by a softmax activation function.
$g(\cdot)$ is used to produce conditional probability $P(Y|X,c_j)$ based on $c_j$.
Inspired by condition injection methods~\cite{huang2017arbitrary,dhariwal2021diffusion}, we design $g$ as follows:
\begin{equation}
\label{eq:f}
\begin{aligned}
    g(f_{img},\mathcal{N}(\mu_{c_j},\sigma^2_{c_j})) 
    &= f_{img}\odot \psi_1(\mu_{c_j}) + \psi_2(\mu_{c_j}), 
\end{aligned}
\end{equation}
where $\odot$ denotes the element-wise product, $\psi_1$ and $\psi_2$ are two independent linear layers.

Equation~\ref{eq:f} only utilizes the mean representation $\mu_{c_j}$ of each clothing item $c_j$ as a condition, which overlooks the variations in clothing representation arise from changes in lighting, viewpoint and other factors in different environments. 
Therefore, we sample $K$ diverse features $[f_{c_{i1}}...f_{c_{iK}}]$ for each clothing item from the distribution $\mathcal{N}(\mu_{c_j},\sigma^2_{c_j})$ to comprehensively account for the influence of clothing conditions. The improved $g(\cdot)$ is represented as follows:
\small
\begin{equation}
\label{eq:f2}
\begin{aligned}
    g(f_{img},\mathcal{N}(\mu_{c_j},\sigma^2_{c_j})) 
    &= \sum_{k=1}^K (f_{img}\odot \psi_1(f_{c_{jk}}) + \psi_2(f_{c_{jk}}))p_{jk}, 
\end{aligned}
\end{equation}
\normalsize
where $p_{jk}$ represents the sampling probabilities of different $f_{c_{ik}}$, which is computed by the probability density function (PDF) of the multivariate Gaussian distribution:
\small
\begin{equation}
\label{eq:pjk}
\begin{aligned}
    &p_{jk} = exp(-\frac{1}{2}D_M(\mu_{c_j},f_{c_{jk}}))/\sum_{k=1}^K exp(-\frac{1}{2}D_M(\mu_{c_j},f_{c_{jk}})), \\
    &D_M(\mu_{c_j},f_{c_{jk}}) = (f_{c_{jk}}-\mu_{c_j})\Sigma^{-1} (f_{c_{jk}}-\mu_{c_j})^T,
\end{aligned}
\end{equation}
\normalsize
where $D_M(\cdot,\cdot)$ is the Mahalanobis distance, $\Sigma\in \mathbb R^{d\times d}$ denotes the diagonal covariance matrix with diagonal elements equal to $\sigma^2_{c_j}$.

To enhance the computational efficiency of Equation~\ref{eq:pydox}, we utilize the Normalized Weighted Geometric Mean (NWGM)~\cite{xu2015show} to approximate moving the classifier out of accumulation operation:
\begin{equation} 
\label{eq:pydox2}
\begin{aligned}     
    P(Y|do(X=x_i)) &= \sum\limits_{c_j} Cls[g(f_{img},\mathcal{N}(\mu_{c_j},\sigma^2_{c_j}))]\cdot P(c_j),\\
    &\approx Cls[\sum\limits_{c_j} g(f_{img},\mathcal{N}(\mu_{c_j},\sigma^2_{c_j}))\cdot P(c_j)].
\end{aligned} 
\end{equation}
This scheme reduces multiple classifications to a single operation, which significantly simplifies the computation of intervention probability.

In conclusion, the Intervention Module can eliminate the influence of the clothing shortcut by simulating causal intervention $P(Y|do(X))$, ensuring clothes-invariant feature learning.

\subsection{Disentangle Regularization} 
\label{sec:disentangle}

To achieve the expectation of modeling the intervention probability $P(Y|do(X))$ by a deep learning implementation, it is crucial to ensure the validity of Equation~\ref{eq:pydox} as much as possible.
Therefore, as shown in Figure~\ref{fig:method} (d), we designed a disentangle regularization to ensure that the clothing representations are discriminative and do not contain identity clues.
This allows for the causal intervention to accurately cut off the clothing shortcut without compromising the extraction of identity features.

We propose a clothes-identity probabilistic orthogonal loss $\mathcal{L}_{po}$ to replace the original classification loss.
Specifically, we construct an ID-Clothes classifier with a total of $N_{clt}+N_{id}$ categories, where $N_{id}$ represents the total human identity categories and $N_{clt}$ denotes the total clothing categories.
Subsequently, the image feature $f_{img}$ and the clothing feature $f_{clt}$ are fed into the ID-Clothes classifier to get the predicted probability distributions:
\begin{equation} 
\begin{aligned}
    p_{img}&=Softmax([W_1;W_2]f_{img}),\\
    p_{cls}&=Softmax([W_2;W_1]f_{clt}),
\end{aligned} 
\end{equation} 
where $W_1 \in \mathcal{R}^{d\times N_{id}}$ and $W_2 \in \mathcal{R}^{d\times N_{clt}}$ are the weights of the classifier, and $[\cdot;\cdot]$ denotes concatenation.
Finally, clothes-identity probabilistic orthogonal loss $\mathcal{L}_{po}$ minimizes the cross-entropy of the predicted probabilities and the identity labels or clothing labels:
\begin{equation} 
\label{eq:po} 
\begin{aligned}
    \mathcal{L}_{po} = \mathbb{E}[-y_i log (p_{img})] + \mathbb{E} [-c_i log (p_{clt})].
\end{aligned} 
\end{equation}

This strategy results in clothes and images becoming negative categories of each other, with their features belonging to different spaces.
Consequently, the clothing features contain distinct information from the image features, facilitating better disentanglement.

To further disentangle clothing and identity, we employ an additional KL-divergence loss $\mathcal{L}_{kl}$:
\begin{equation}
\label{eq:po2}
\begin{aligned}
    \mathcal{L}_{kl} = \mathbb{E}[D_{kl}(P_{img}||P'_{img}) + D_{kl}(P_{clt}||P'_{clt})],
\end{aligned}
\end{equation}
where $D_{kl}(\cdot||\cdot)$ denotes the KL divergence. $P_{img}$ is computed by averaging the classification probabilities $p_{img}$ from half of the samples with the same identity label in the training batch, while $P'_{img}$ denotes the average classification probabilities from the other half of the samples.
$P_{clt}$ and $P'_{clt}$ has a similar meaning and calculation pipeline.
So this loss term means that the classification probabilities among positive samples should be as similar as possible and reduce the interference of noisy samples through probability averaging.

In addition, we design a metric learning loss, clothes-identity feature separation loss $\mathcal{L}_{fs}$, that directly constrains the Euclidean distance between features: \begin{equation} 
\label{eq:fo} 
\begin{aligned}     
\mathcal{L}_{fs} &= \mathbb{E} [D_{eu}(f^a_m,f^p_m)] + \mathbb{E} [\rho-D_{eu}(f^a_m,f^n_m)]_+,
\end{aligned} 
\end{equation} 
where $D_{eu}(\cdot,\cdot)$ is Euclidean distance, $[\cdot]_+=max(0,\cdot)$, $\rho$ is the margin parameter.
And $m\in\{img,clt\}$, $f_m$ represents the image/clothing feature centers from half of the samples with the same identity/clothing label in the current mini-batch. 
$f^p$ is the feature center of the positive samples for $f^a$ and $f^n$ is the negative one.

Specifically, this loss aims to cluster positive features closely together and distance negative features from each other.
Similar to our proposed $L_{po}$ loss, the $L_{fs}$ loss conducts feature learning in a shared metric space, where clothes and images serve as negative categories for each other, emphasizing their distinction. 
This strategy further enhances that the clothes features should have different information from the image ones, leading to better disentanglement. 

\subsection{Optimization}
\label{sec:opt}
The whole model is trained end-to-end and the total loss $\mathcal{L}_{total}$ of our method is defined as:
\begin{equation}
\begin{aligned}
\label{eq:opt}
\mathcal{L}_{total}=\mathcal{L}_{int}+\mathcal{L}_{po}+\mathcal{L}_{kl}+\mathcal{L}_{fs},
\end{aligned}
\end{equation}
We uniformly assigned a weight of 1 to all losses, without the need for hyper-parameter searches, which has yielded good results.

In addition, considering that the Confounder Dictionary may not be accurate in the early stages of training, we do not apply the intervention loss $L_{int}$ during the warmup stage (first 10 epochs) to stabilize the training.
 
\section{Experiments}
\label{sec:exper}
We focus on evaluating the CC-ReID performance of CCIL under multiple scenarios, verifying the effectiveness of causal intervention in suppressing the clothing shortcut, and assessing the robustness of CCIL under noisy clothing labels. The Supplementary Material reports additional analyses on complexity, occlusion, and other settings.

\subsection{Datasets and Evaluation Protocol}

{\bf Dataset Details.}
For primary evaluation, we use six widely used public CC-ReID datasets: PRCC~\cite{yang2019person}, VC-Clothes~\cite{wan2020person}, LTCC~\cite{qian2020long}, DeepChange~\cite{xu2023deepchange}, Celeb-reID-light~\cite{huang2019beyond} and LaST~\cite{shu2021large}.
Table~\ref{tab:dataset} gives brief statistics of the datasets used in this work.
The \textbf{PRCC} dataset is collected from 3 cameras and each identity has 2 pieces of clothing. It contains 33,698 images from 221 identities.
The \textbf{VC-Clothes} dataset is a synthetic CC-ReID dataset rendered by the GTA5 game engine.
It contains 512 virtual identities of 19,060 images in 4 different cameras.
The \textbf{LTCC} dataset contains 17,119 images of 152 identities captured by 12 cameras.
The \textbf{DeepChange} dataset is a large-scale long-term ReID dataset, which consists of 178,407 images of 1,121 identities from 17 cameras and only includes the recording date without clothing labels.
The \textbf{LaST} dataset is also a long-term ReID benchmark collected from more than 2,000 movies in 8 countries, containing 10,862 identities and 228,156 images.
The \textbf{Celeb-reID-light} dataset is collected from snapshots of celebrities on the Internet. It contains 10,842 images of 590 identities. In this dataset, each clothing item has only one image, so image IDs can be used as clothing IDs.

{\bf Evaluation Protocol.}
Our experiments follow the evaluation protocol in existing CC-ReID benchmarks.
The Rank-k accuracy and mean average precision (mAP) are adopted as the evaluation metrics.
We employ a \textbf{cloth-changing} evaluation setting for PRCC, VC-Clothes, and LTCC datasets. 
In this setting, the clothing between each individual's query and gallery images is different.
We also report the result under the \textbf{cloth-unchanging} setting for PRCC and VC-Clothes datasets, which means images are all cloth-consistent for each identity.
For the PRCC dataset, we follow~\cite{huang2021clothing,gu2022clothes} and report results on cameras A/C for cloth-changing and A/B for cloth-unchanging settings.
For the VC-Clothes dataset, we follow~\cite{huang2021clothing,gu2022clothes} and report results on cameras 3/4 for cloth-changing and 2/3 for cloth-unchanging settings.
For the LTCC dataset, the accuracy is calculated only using cloth-changing ground-truth samples in the cloth-changing setting~\cite{hong2021fine,yang2023good}.
For the DeepChange dataset, we follow~\cite{xu2023deepchange,gu2022clothes} and use true matches from different times and trajectories.
For Celeb-reID-light and LaST, we follow their official standard ReID evaluation protocols.

\begin{table}[!t]
    \footnotesize
    \setlength\tabcolsep{3.0pt}
    \renewcommand{\arraystretch}{1.1}
    \centering
    \caption{Brief statistics of datasets used in this work.}
    \label{tab:dataset}
    \begin{tabular}{l l r r c c}
        \toprule[1pt]
        Dataset & Source & Images & IDs & Cameras & Cloth-Labels
        \cr\midrule
        PRCC & Surveillance & 33,698 & 221 & 3 & \Checkmark\cr
        VC-Clothes & Synthetic & 19,060 & 512 & 4 & \Checkmark\cr
        LTCC & Surveillance & 17,119 & 152 & 12 & \Checkmark\cr
        DeepChange & Surveillance & 178,407 & 1,121 & 17 & \XSolidBrush\cr
        LaST & Movie & 228,156 & 10,862 & - & \Checkmark\cr
        Celeb-reID-light & Internet & 10,842 & 590 & - & \Checkmark\cr 
        \bottomrule[1pt]
    \end{tabular}
    \vspace{-3mm}
\end{table}

\subsection{Implementation Details}

{\bf Network.}
We utilize the ResNet-50 model~\cite{he2016deep} as the backbone of our two-stream network.
Following the widely used Re-ID methods~\cite{luo2019bag} the last convolutional stride is set to 1 and the BNNeck is added.
The first bottleneck of ResNet is designated as the shared module, while other bottlenecks are configured as branch-specific modules.
We also employed the ViT-Base model~\cite{dosovitskiy2020image} pre-trained on ImageNet with patch size 16, as the backbone of our two-stream network to further test the generalizability of our method.
The first four layers of ViT are designated as shared modules, while the other layers are configured as branch-specific modules.

{\bf Training.}
The model is trained for 120 epochs with the SGD optimizer.
In the first 10 epochs, the learning rate linearly increased from 0.001 to 0.01 for the ResNet-50 backbone and increased from 0.0008 to 0.008 for the ViT backbone. 
Afterwards, the learning rate was decayed to 0 following a cosine decay schedule.
The batch size is set to 64 with 8 identities and 1 or 2 clothing items.
Following~\cite{qian2020long,gu2022clothes,yang2023good,han2023clothing}, images are resized to 384$\times$192 and are augmented with random horizontal flipping, padding, random cropping, and random erasing~\cite{zhong2020random} in training.
Following~\cite{qian2020long,chen2021learning,cui2023dcr,gu2022clothes,yang2023good,han2023clothing}, clothing labels are used during training.
The hyper-parameter $\alpha$ in Equation~\ref{eq:alpha} is set to 0.9, $K$ in Equation~\ref{eq:f2} is set to 4, and $\rho$ in Equation~\ref {eq:fo} is set to 0.6 for the ResNet-50 backbone and set to 0.9 for the ViT backbone.

{\bf Inference.}
In the inference stage, only the identity feature extraction stream is activated to extract $f_{img}$ as a human signature, and we directly measure cosine similarities across images to obtain the retrieval results.
The clothing stream and causal modules are used only for training, so their additional cost does not affect inference; detailed parameter and runtime comparisons are provided in the Supplementary Material.

\subsection{Comparison with Causality-based ReID Methods}
We compared CCIL with causality-based ReID methods~\cite{rao2021counterfactual,yang2023good} on the PRCC dataset. Following~\cite{lopez2017discovering, wang2020visual}, in addition to common accuracy metrics, we introduced the neural causation coefficient (NCC) to quantitatively verify the effectiveness of approaches in eliminating clothing shortcuts.
NCC is a causal discovery model pretrained on synthetic observation samples that can directly operate on feature vectors to evaluate causal relationships in visual images.
Specifically, we input identity features and clothing features into the pre-trained NCC model to obtain the NCC($C\rightarrow X$) score, which ranges from (0,1) and represents the relative causality intensity from clothes $C$ to image $X$.
Due to factors such as blurriness, occlusion, and variations in lighting, some samples exhibit insignificant clothing shortcuts prior to intervention.
Therefore, we report the average values of the NCC($C\rightarrow X$) score for the top 10\% samples to highlight the methods' effectiveness in mitigating clothing shortcuts.
Causal interventions can cut off the clothing shortcut by disrupting the potential correlation between the intervention variable $X$ and its cause $C$, as illustrated by the removed arrow $C \rightarrow X$. 
Consequently, a lower NCC($C \rightarrow X$) indicates a more effective prevention of interference caused by clothing shortcuts.

As shown in Table~\ref{tab:ncc}, our method achieves the best NCC scores and Rank-1 accuracy, significantly outperforming the baseline and other causality-based ReID methods.
This suggests that our approach effectively eliminates clothing shortcuts, while other causality-based ReID methods fail to do so.
Method~\cite{rao2021counterfactual} employs counterfactual interventions to optimize the total direct effect (TDE) of attention maps. 
This approach attempts to eliminate all potential spurious correlations in spatial attention without specifically addressing the influence of clothing, thus failing to effectively eliminate clothing shortcuts.
TDE-based methods, such as AIM~\cite{yang2023good}, highlight clothes-unrelated features within individual images through factual-counterfactual comparison. However, they do not explicitly adjust the clothing distribution associated with each image. In contrast, for each clothing item, backdoor adjustment blocks the shortcut path $X\leftarrow C\rightarrow Y$ by replacing $P(C|X)$ with $P(C)$. Since clothing shortcuts in CC-ReID mainly stem from distribution-level clothing-identity spurious correlations in the training data, this global adjustment directly addresses the source of the bias and therefore suppresses such shortcuts in a more targeted manner.

\begin{table}[!t]
    \setlength\tabcolsep{10.0pt}
    \renewcommand{\arraystretch}{1.1}
    \centering
    \caption{
    Comparison with causality-based ReID methods on the PRCC dataset. The neural causation coefficient (NCC) is used to quantitatively verify the effectiveness of approaches in eliminating clothing shortcuts. The terms "TDE" and "BA" refer to total direct effect and backdoor adjustment.
    }
    \label{tab:ncc}
    \begin{tabular}{l l c c}
        \toprule[1pt]
        Method&Causal Technology&
        NCC$\downarrow$&Rank1$\uparrow$
        \cr\midrule
        CAL~\cite{rao2021counterfactual}&
        TDE (Counterfactual)&
        0.32&55.0\cr
        AIM~\cite{yang2023good}&
        TDE (Counterfactual)&
        0.26&57.9\cr
        \midrule
        baseline&
        No Causal&
        {0.34}&{54.1}\cr
        {CCIL (Ours)}&BA (Intervention)&
        {0.13}&{66.4}\cr
        \bottomrule[1pt]
    \end{tabular}
    \vspace{-3mm}
\end{table}

\begin{table*}[!t]
    \setlength\tabcolsep{4.0pt}
    \renewcommand{\arraystretch}{1.1}
    \centering
    \caption{
    Comparison with the state-of-the-art methods on three CC-ReID datasets. The terms "pose", "2D", and "3D" refer to human poses, 2D silhouettes, and 3D shape information, respectively. ''-'' denotes that the original paper was not reported.
    }
    \label{tab:comp}
    \begin{tabular}{l l l c c c c c c c c c c}
        \toprule[1pt]
        \multirow{3}*[-0.8em]{Method}&
        \multirow{3}*[-0.8em]{Venue}&
        \multirow{3}*[-0.8em]{Modality}&
        \multicolumn{4}{c}{PRCC}&
        \multicolumn{4}{c}{VC-Clothes}&
        \multicolumn{2}{c}{LTCC}
        \cr\cmidrule(r){4-7}\cmidrule(r){8-11}\cmidrule(r){12-13}&&&
        \multicolumn{2}{c}{Cloth-changing}&         \multicolumn{2}{c}{Cloth-unchanging}&         \multicolumn{2}{c}{Cloth-changing}&         \multicolumn{2}{c}{Cloth-unchanging}&
        \multicolumn{2}{c}{Cloth-changing}\cr\cmidrule(r){4-5}\cmidrule(r){6-7}\cmidrule(r){8-9}\cmidrule(r){10-11}\cmidrule(r){12-13}
        &&&\cellcolor[HTML]{FFFFFF}Rank1&\cellcolor[HTML]{FFFFFF}mAP&Rank1&mAP&\cellcolor[HTML]{FFFFFF}Rank1&\cellcolor[HTML]{FFFFFF}mAP&Rank1&mAP&\cellcolor[HTML]{FFFFFF}Rank1&\cellcolor[HTML]{FFFFFF}mAP
        \cr\midrule
        SPT+ASE~\cite{yang2019person}&TPAMI'2019&Contour
        &34.4&-&64.2&-&-&-&-&-&-&-\cr
        GI-ReID~\cite{jin2022cloth}&CVPR'2022&RGB+2D
        &37.6&-&80.0&-&64.5&57.8&-&-&23.7&10.4\cr
        UCAD~\cite{yan2022weakening}&IJCAI'2022&RGB+2D
        &45.3&-&96.5&-&82.4&73.8&92.6&81.1&32.5&15.1\cr
        3DSL~\cite{chen2021learning}&CVPR'2021&RGB+Pose+2D+3D
        &51.3&-&-&-&79.9&81.2&-&-&31.2&14.8\cr
        FSAM~\cite{hong2021fine}&CVPR'2021&RGB+Pose+2D
        &54.5&-&98.8&-&78.6&78.9&94.7&94.8&38.5&16.2\cr
        DCR-ReID~\cite{cui2023dcr}&TCSVT'2023&RGB+Parsing+Contour
        &57.2&57.4&\textbf{100.0}&99.7&-&-&-&-&41.1&20.4\cr
        CCPG~\cite{nguyen2024contrastive}&CVPR'2024&RGB+2D
        &61.8&58.3&\textbf{100.0}&99.6&-&-&-&-&\textbf{46.2}&\textbf{22.9}\cr
        HPRNet~\cite{xiong2025hprnet}&TCSVT'2025&RGB+Parsing
        &62.3&60.1&\textbf{100.0}&99.6&-&-&-&-&45.9&19.2\cr
        \midrule
        IANet~\cite{hou2019interaction}&CVPR'2019&RGB    &46.3&46.9&99.4&98.3&-&-&-&-&25.0&12.6\cr
        mAPLoss~\cite{shu2021large}&TCSVT'2021&RGB
        &57.5&54.7&-&-&-&-&-&-&-&-\cr
        CAL~\cite{gu2022clothes}&CVPR'2022&RGB         &55.2&55.8&\textbf{100.0}&\underline{99.8}&81.4&81.7&\underline{95.1}&\underline{95.3}&40.1&18.0\cr
        ACID~\cite{yang2023win}&TIP'2023&RGB
        &55.4&-&99.1&-&84.3&74.2&\underline{95.1}&94.7&29.1&14.5\cr

        AIM~\cite{yang2023good}&CVPR'2023&RGB
        &57.9&58.3&\textbf{100.0}&\textbf{99.9}&82.1&\underline{81.9}&95.0&95.1&40.6&19.1\cr
        CCFA~\cite{han2023clothing}&CVPR'2023&RGB
        &61.2&58.4&99.6&98.7&-&-&-&-&45.3&22.1\cr
        Instruct-ReID~\cite{he2024instruct}&CVPR'2024&RGB
        &54.2&52.3&-&-&\textbf{89.7}&78.9&-&-&-&-\cr
        FIRe$^2$~\cite{wang2024exploring}&TIFS'2024&RGB
        &\underline{65.0}&\underline{63.1}&\textbf{100.0}&99.5&-&-&-&-&44.6&19.1\cr

        \midrule
        {CCIL}&{Ours}&RGB
        &{\textbf{66.4}}&{\textbf{65.2}}
        &{\textbf{100.0}}&99.2
        &{\underline{89.6}}&{\textbf{88.2}}
        &{\textbf{96.0}}&{\textbf{95.7}}
        &{\underline{46.0}}&{\underline{22.2}}\cr
        \bottomrule[1pt]
    \end{tabular}
    \vspace{-3mm}
\end{table*}

\begin{table}[!t]
    \setlength\tabcolsep{8.0pt}
    \renewcommand{\arraystretch}{1.1}
    \centering
    \caption{Comparison with the state-of-the-art methods on the DeepChange dataset without clothes labels.}
    \label{tab:comp2}
    \begin{tabular}{l c c c c}
        \toprule[1pt]
        \multirow{2}*[-0.4em]{Method}&
        \multicolumn{4}{c}{DeepChange}
        \cr\cmidrule(r){2-5}&
        Rank1&Rank5&Rank10&mAP
        \cr\midrule
        BoT ResNet-50~\cite{luo2019strong}&
        47.5&59.5&65.2&13.0\cr
        ReIDCaps~\cite{huang2019beyond}&
        44.3&56.4&62.0&13.3\cr
        ViT B16~\cite{dosovitskiy2020image}&
        49.7&61.8&67.4&15.0\cr
        SCNet~\cite{guo2023semantic}&
        53.5&-&-&18.7\cr
        CAL~\cite{gu2022clothes}&
        54.0&-&-&\underline{19.0}\cr
        IMS+GEP~\cite{zhao2023joint}&
        \underline{55.1}&\underline{64.9}&\underline{69.6}&18.3\cr
        baseline (Ours)&53.6&64.2&69.3&17.2\cr\hline
        {CCIL (Ours)}&
        {\textbf{59.2}}&{\textbf{69.5}}&{\textbf{74.3}}&{\textbf{20.8}}
        \cr\bottomrule[1pt]
    \end{tabular}
    \vspace{-3mm}
\end{table}

\begin{table}[!t]
    \setlength\tabcolsep{4.2pt}
    \renewcommand{\arraystretch}{1.1}
    \centering
    \begingroup
    \caption{Comparison with the state-of-the-art methods on the LaST and Celeb-reID-light datasets.}
    \label{tab:comp5}
    \begin{tabular}{l c c l c c}
        \toprule[1pt]
        \multirow{2}*[-0.4em]{Method}&
        \multicolumn{2}{c}{LaST}&
        \multirow{2}*[-0.4em]{Method}&
        \multicolumn{2}{c}{Celeb-reID-light}
        \cr\cmidrule(r){2-3}\cmidrule(r){5-6}
        &Rank1&mAP&&Rank1&mAP
        \cr\midrule
        mAPLoss~\cite{shu2021large}&71.0&28.0&
        mAPLoss~\cite{shu2021large}&29.0&16.3\cr

        IMS+GEP~\cite{zhao2023joint}&73.2&29.8&
        RCSANet~\cite{huang2021clothing}&29.3&16.7\cr
        
        CAL~\cite{gu2022clothes}&73.7&28.8&
        CAL~\cite{gu2022clothes}&33.6&18.5\cr

        FIRe$^2$~\cite{wang2024exploring}&\underline{75.0}&\textbf{32.2}&
        3DInvarReID~\cite{liu2023learning}&\underline{37.0}&\underline{21.8}\cr

        baseline (ours)&72.8&27.9&
        baseline (ours)&31.5&17.0\cr
        \midrule
        {CCIL (Ours)}&\textbf{76.8}&\underline{32.0}&
        {CCIL (Ours)}&\textbf{38.5}&\textbf{22.4}
        \cr\bottomrule[1pt]
    \end{tabular}
    \endgroup
    \vspace{-3mm}
\end{table}

\subsection{Comparison with State-of-the-art Methods}

We compare the performance of CCIL with state-of-the-art methods on PRCC, VC-Clothes, LTCC, DeepChange, LaST and Celeb-reID-light datasets in Table~\ref{tab:comp}, Table~\ref{tab:comp2} and Table~\ref{tab:comp5}.
The PRCC dataset provides standardized evaluation, while VC-Clothes, LTCC, and DeepChange datasets respectively consider the effectiveness of the methods under the conditions of clothing sharing, some pedestrians not changing outfits, and the absence of clothing labels. LaST and Celeb-reID-light further evaluate generalization in non-surveillance scenarios.

{\bf Results on Ideal Conditions.} PRCC is a standard CC-ReID dataset, in which each identity is associated with two sets of clothing and appears in all cameras, making it well-suited for investigating the pure impact of clothing.
As depicted in Table~\ref{tab:comp}, CCIL achieves 66.4\% Rank-1 accuracy and 65.2\% mAP accuracy on the PRCC dataset.
Compared with the methods using auxiliary modality data, our method surpasses HPRNet~\cite{xiong2025hprnet} by a large margin, with 4.1\% absolute improvement in Rank-1 accuracy, along with 5.1\% enhancements in mAP accuracy.
Although these existing methods incorporate clothes-invariant modality information such as poses and contours, most of them use the RGB modality as the main input, resulting in the persistence of clothing shortcuts that interfere with feature learning.
Our method achieves superior results without the need for additional modality data.
Compared with the methods using RGB modality only, our method outperforms the second-best method FIRe$^2$~\cite{wang2024exploring} by 1.4\% and 2.1\% in Rank-1 and mAP accuracy.
From our results, it can be seen that it is still challenging for the likelihood-based methods to drop clothes-related cues while keeping the identity discriminative, further proving the effectiveness of our causality-based work.

Furthermore, under the cloth-unchanging setting, our approach achieves the best Rank-1 and comparable mAP accuracy among all the state-of-the-art methods.
The cloth-unchanging setting is based on the strong assumption that people keep their clothes unchanged, allowing the utilization of clothing information to bring gains.
Our method aims to capture reliable identity cues while excluding unreliable clothing information, enabling generalization across diverse scenarios, regardless of whether the clothing is changed.

{\bf Results with Shared Clothing Conditions.} VC-Clothes is a synthetic CC-ReID dataset in which a small number of clothing items are shared, resembling the low-probability scenario in the real world where two individuals wear the same outfit.
Since most clothing is not shared, the clothing shortcuts for inferring identity based on clothing will still significantly interfere with the learning of clothes-invariant features.
Furthermore, shared clothing does not affect the effectiveness of the Confounder Dictionary and the Intervention Module, as it equates to storing multiple representations for a clothing item.
As shown in Table~\ref{tab:comp}, CCIL achieves the best mAP and Rank-1 accuracy among all the state-of-the-art competitors on the VC-Clothes dataset under both the cloth-changing and cloth-unchanging settings.
The results indicate that our method remains effective in the presence of clothing sharing. 
We also point out that if the majority of clothing is shared among most identities, the clothing shortcuts become very weak, although this is unlikely to occur. 
In this extreme case, $P(Y|do(X))$ remains the correct objective, but the improvement effect is limited, as $P(Y|X)$ is already very close to $P(Y|do(X))$.

{\bf Results with Limited Clothing Diversity.} LTCC is a small CC-ReID dataset that contains a total of 77 identities in the training set, with 31 identities having only one set of clothing.
As depicted in Table~\ref{tab:comp}, CCIL achieves better or comparable performance compared to FIRe$^2$~\cite{wang2024exploring} and CCFA~\cite{han2023clothing} on the LTCC dataset.
Compared with AIM~\cite{yang2023good}, CCIL also achieves higher Rank-1 and mAP accuracy on the Cloth-changing setting of PRCC.
We further illustrate the reasons for the relatively fewer improvements on the LTCC dataset.
Since LTCC only includes 46 identities with multiple sets of clothing, which provides fewer clothing items, the modeling quality of our Confounder Dictionary may be somewhat affected.

{\bf Results without Clothing Labels.} DeepChange is a large-scale CC-ReID dataset that does not provide clothing labels.
Following~\cite{gu2022clothes,yang2023good}, we use the date of shooting as pseudo clothing labels.
The results on DeepChange shown in Table~\ref{tab:comp2} further show the superiority of our method.
Our method enhanced the baseline in Rank-1 accuracy and mAP accuracy by 5.6\% and 3.6\%, respectively, and outperformed IMS+GEP by 4.1\% and 2.5\% in Rank-1 accuracy and mAP accuracy.

{\bf Results on Non-surveillance Scenarios.} 
To further evaluate generalization beyond surveillance scenarios, we conduct experiments on LaST (movie scenes) and Celeb-reID-light (web images). 
As shown in Table~\ref{tab:comp5}, compared with the baseline, CCIL improves Rank-1 by 4.0\% on LaST and 7.0\% on Celeb-reID-light.
Compared with previous CNN-based methods, CCIL achieves the best Rank-1 on both datasets, with gains of 1.8\% on LaST and 1.5\% on Celeb-reID-light.

{\bf Results with Transformer Architecture.}
As depicted in Table~\ref{tab:comp3}, we combine CCIL with transformer architecture to evaluate the generalizability of our approach.
When using ViT-Base instead of ResNet-50 as the backbone network, our method achieves 71.2\% and 91.0\% rank-1 accuracies on PRCC and VC-Clothes, respectively, surpassing competitive transformer-based methods~\cite{wei2025multiple,zhang2025adaptive}.
Our CCIL does not modify the backbone network, making it model-agnostic and compatible with various architectures.

\begin{table}[!t]
    \renewcommand{\arraystretch}{1.1}
    \centering
    \caption{Comparison with the state-of-the-art methods using Transformer (ViT~\cite{dosovitskiy2020image}) backbone on the Cloth-changing setting of PRCC and VC-Clothes datasets.}
    \label{tab:comp3}
    \begin{tabular}{l l c c c c}
        \toprule[1pt]
        \multirow{2}*[-0.4em]{Method}&
        \multirow{2}*[-0.4em]{Venue}&
        \multicolumn{2}{c}{PRCC}&
        \multicolumn{2}{c}{VC-Clothes}
        \cr\cmidrule(r){3-4}\cmidrule(r){5-6}&&
        Rank1&mAP&Rank1&mAP
        \cr\midrule
        MIPL~\cite{wei2025multiple}&TIP'2025&
        69.2&64.8&-&-\cr
        A$^3$PFN~\cite{zhang2025adaptive}&PR'2025&
        69.1&68.7&89.2&83.1\cr
        \midrule
        CCIL (ViT)&Ours&
        71.2&66.8&91.0&86.4
        \cr\bottomrule[1pt]
    \end{tabular}
    \vspace{-3mm}
\end{table}

\subsection{Ablation Study}

In this section, we conduct ablation studies on several CC-ReID datasets to evaluate the effectiveness of each detailed part of the CCIL, including the Intervention Module, the Confounder Dictionary, and the Disentangle Regularization. 

\begin{table}[!t]
    \setlength\tabcolsep{4.2pt}
    \renewcommand{\arraystretch}{1.1}
    \centering
    \caption{Ablation experiments of the proposed Intervention Module on the Cloth-changing setting of PRCC and LTCC datasets.}
    \label{tab:intervention}
    \begin{tabular}{c l c c c c}
        \toprule[1pt]
        \multirow{2}*[-0.4em]{Index}&
        \multirow{2}*[-0.4em]{Intervention Module}&
        \multicolumn{2}{c}{PRCC}&
        \multicolumn{2}{c}{LTCC}
        \cr\cmidrule(r){3-4}\cmidrule(r){5-6}&&
        Rank1&mAP&Rank1&mAP
        \cr\midrule
        1&No Causal Intervention
        &54.1&54.7&34.3&15.3\cr\midrule
        2&Addition
        &57.2&56.6&36.9&17.6\cr  
        3&Concatenation
        &56.3&55.8&36.2&17.0\cr  
        4&Cross Attention
        &58.3&57.5&38.3&18.2\cr\midrule
        5&Eq.~\ref{eq:f} (Ours w/o distribution)                  
        &59.4&58.7&38.8&18.4\cr
        6&Eq.~\ref{eq:f2} (Ours)         
        &61.0&59.8&40.6&19.0\cr
        \bottomrule[1pt]
    \end{tabular}
    \vspace{-3mm}
\end{table}

\begin{table}[!t]
    \setlength\tabcolsep{3.2pt}
    \renewcommand{\arraystretch}{1.1}
    \centering
    \caption{Ablation experiments of the proposed Confounder Dictionary on the Cloth-changing setting of PRCC and DeepChange datasets.}
    \label{tab:dictionary}
    \begin{tabular}{c l c c c c}
        \toprule[1pt]
        \multirow{2}*[-0.4em]{Index}&
        \multirow{2}*[-0.4em]{Confounder Dictionary}&
        \multicolumn{2}{c}{PRCC}&
        \multicolumn{2}{c}{DeepChange}
        \cr\cmidrule(r){3-4}\cmidrule(r){5-6}&&
        Rank1&mAP&Rank1&mAP
        \cr\midrule
        1&Dictionary$\rightarrow$Random&
        54.9&55.2&53.1&17.2\cr         
        2&Dictionary$\rightarrow$Learnable&
        54.2&54.8&53.3&17.3\cr\midrule
        3&Dictionary w/o $M_{clt}$         
        &59.3&58.0&55.3&18.0\cr         
        4&Dictionary w/o $1-M_{img}$         
        &60.0&58.5&55.8&18.3\cr\midrule
        5&Dictionary$\rightarrow$All $f_{clt}$ (Random)&
        58.7&58.0&54.8&17.9\cr
        6&Dictionary$\rightarrow$All $f_{clt}$ (Average)&
        59.1&58.2&55.1&18.1\cr\midrule
        7&Dictionary (Ours)&
        61.0&59.8&56.1&18.8\cr
        \bottomrule[1pt]
    \end{tabular}
    \vspace{-3mm}
\end{table}

\begin{table}[!t]
    \setlength\tabcolsep{4.8pt}
    \renewcommand{\arraystretch}{1.1}
    \centering
    \caption{Ablation experiments of the proposed Disentangle Regularization on the Cloth-changing setting of PRCC and VC-Clothes datasets.}
    \label{tab:disentangle}
    \begin{tabular}{c c c c c c c c c}
        \toprule[1pt]
        \multirow{2}*[-0.4em]{Index}&
        \multirow{2}*[-0.4em]{$L_{int}$}&
        \multicolumn{3}{c}{Disentangle}&
        \multicolumn{2}{c}{PRCC}&
        \multicolumn{2}{c}{VC-Clothes}
        \cr\cmidrule(r){3-5}\cmidrule(r){6-7}\cmidrule(r){8-9}
        &&$L_{po}$&$L_{kl}$&$L_{fs}$&
        Rank1&mAP&Rank1&mAP
        \cr\midrule
        1&\Checkmark&--&--&--&  61.0&59.8&84.7&83.6\cr
        2&\Checkmark&\Checkmark&--&--&  61.9&61.0&86.3&85.7\cr
        3&\Checkmark&--&\Checkmark&--&  63.6&62.8&87.3&86.5\cr
        4&\Checkmark&--&--&\Checkmark&  64.1&63.3&87.5&86.6\cr
        5&\Checkmark&\Checkmark&\Checkmark&\Checkmark&  66.4&65.2&89.6&88.2\cr
        6&--&\Checkmark&\Checkmark&\Checkmark&  61.4&60.5&84.2&82.6\cr
        \bottomrule[1pt]
    \end{tabular}
    \vspace{-3mm}
\end{table}

{\bf Effectiveness of the Intervention Module.}
As shown in Table~\ref{tab:intervention}, in the 1-$st$ row, we establish a baseline, which is a two-stream network without any causal modules and trained by the standard cross-entropy loss for image features and clothing features.
In the 2$^{nd}$ $\sim$ 6$^{th}$ rows, we introduce causal intervention by different implementations of the backdoor adjustment corresponding to Equation~\ref{eq:pydox}.
Specifically, the 2$^{nd}$ $\sim$ 4$^{th}$ rows represent the fusion of features for $f_{img}$ and $f_{c_j}$ through addition~\cite{nan2021interventional}, concatenation~\cite{deng2021comprehensive,liu2022show,zhang2022learning}, and cross-attention~\cite{wang2020visual,zhang2022multiple,li2021interventional,lin2022causal,lin2023interventional,huang2023causalainer} to achieve $P(Y|do(X))$, while the 5$^{th}$ and 6$^{th}$ rows illustrate our non-distributed implementation in Equation~\ref{eq:f} and the distributed implementation Equation~\ref{eq:f2}.
It is evident that each of these different realizations significantly enhances the baseline, illustrating the advantages of pursuing pure causal effects as opposed to solely focusing on correlations.
Our intervention module achieved optimal performance, surpassing widely used realizations such as the cross-attention mechanism.
Our distributed version further enhanced the effectiveness of the intervention, as it is equivalent to using a larger and more comprehensive Confounder Dictionary.
Ultimately, compared to the baseline, our implementation achieved 6.9\% and 6.3\% improvement in Rank-1 accuracy and 5.1\% and 4.7\% improvement in mAP accuracy on the PRCC and LTCC datasets.

\begin{table}[!t]
    \setlength\tabcolsep{4.2pt}
    \renewcommand{\arraystretch}{1.1}
    \centering
    \begingroup
    \caption{Robustness to clothing label quality on the Cloth-changing setting of PRCC.}
    \label{tab:label_robust}
    \begin{tabular}{c l c c}
        \toprule[1pt]
        Index & Clothing Label Setting & Rank1 & mAP \\
        \midrule
        1&No Causal Intervention & 54.1 &54.7 \\
        \midrule
        2&50\% Within-ID Clothing-Label Randomization & 60.4 & 59.2 \\
        3&100\% Within-ID Clothing-Label Randomization & 59.2 & 58.1 \\
        4&ID Labels & 59.3 & 58.3 \\
        5&Clustering Pseudo Labels & 60.5 & 59.4 \\
        \midrule
        6&Clothing Labels & 61.0 & 59.8 \\
        \bottomrule[1pt]
    \end{tabular}
    \endgroup
    \vspace{-3mm}
\end{table}

{\bf Effectiveness of the Confounder Dictionary.}
To further prove that the reason for the performance gain is causality rather than introducing other parameters, we conduct a series of further experiments in Table~\ref{tab:dictionary}.
In the 1$^{st}$ row, we use randomly initialized vectors as the values in the Confounder Dictionary and keep them fixed.
In the 2$^{nd}$ row, we instead treat the Confounder Dictionary as learnable parameters.
The purpose of this experiment is to show what happens when we remove the causal meaning of intervention but keep the training scheme unchanged, treating it as a feature enhancement without causal meaning.
The results show that they bring average performance drops of 6.1\% and 3.0\% in Rank-1 accuracy and 4.6\% and 1.6\% in mAP accuracy on the PRCC and DeepChange datasets compared with our causal intervention method.
When the Confounder Dictionary lacks valid constraints for deriving clothing representations, the optimization objective of the Intervention Module deviates from causal intervention.
In contrast, the results of the 3$^{rd}$ and 4$^{th}$ rows demonstrate that incorporating spatial information to effectively constrain clothing representations is beneficial, as more accurate confounder modeling allows for more effective causal intervention.
Furthermore, in the 5$^{th}$ and 6$^{th}$ rows, we attempt to directly store the clothing features $f_{clt}$ of all images, which resulted in a 25-fold increase in the size of the dictionary on the PRCC dataset.
We employ random sampling (5$^{th}$ row) or averaging (6$^{th}$ row) for each garment to avoid the substantial time costs caused by intervening on each $f_{clt}$.
The results indicate that our multivariate Gaussian distribution modeling method can generate more diverse representations for each garment, leading to the best performance.

{\bf Effectiveness of the Disentangle Regularization.}
The experiments conducted in Table~\ref{tab:disentangle} explore the effectiveness of the Disentangle Regularization.
Rows~2$\sim$4 evaluate each disentanglement loss independently under the same causal-intervention framework. Compared with the baseline using only $L_{int}$, adding $L_{po}$, $L_{kl}$, and $L_{fs}$ separately improves the Rank-1 accuracy on PRCC by 0.9\%, 2.6\%, and 3.1\%, respectively.
Specifically, our proposed $L_{po}$ loss encourages probabilistic orthogonality between identity and clothing categories, $L_{kl}$ stabilizes the predicted distributions among positive samples, while the $L_{fs}$ loss separates the clothing and identity features, further supporting feature disentanglement.
The full combination achieves the best performance on both datasets, demonstrating their complementarity.
With the Disentangle Regularization, the network can learn pure clothing features, which do not include identity-related information. 
Consequently, the Confounder Dictionary can better model the clothing $C$, and the Intervention Module can also better estimate $P(Y|do(X))$ by a neural network.
The setting using all three disentanglement losses without $L_{int}$ still performs worse than the full model.
This demonstrates that disentanglement cannot replace causal intervention.
All these ablation experiments demonstrate the effectiveness of our CCIL design for achieving clothes-invariant features.

\subsection{Robustness of Causal Intervention to Clothing Labels}
In Section V-D, we use shooting dates as pseudo clothing labels on DeepChange. In this section, we further analyze the sensitivity of causal intervention to clothing-label quality. 
The evaluated settings include within-ID clothing-label randomization at different ratios and generating pseudo clothing labels via clustering of clothing features.
Since clothing labels in CC-ReID are defined per identity (distinguishing different outfits of the same person), all random relabeling and clustering operations are conducted independently within each identity, without cross-identity mixing.

\begin{figure}[!t]
    \centering
    \includegraphics[width=0.90\linewidth]{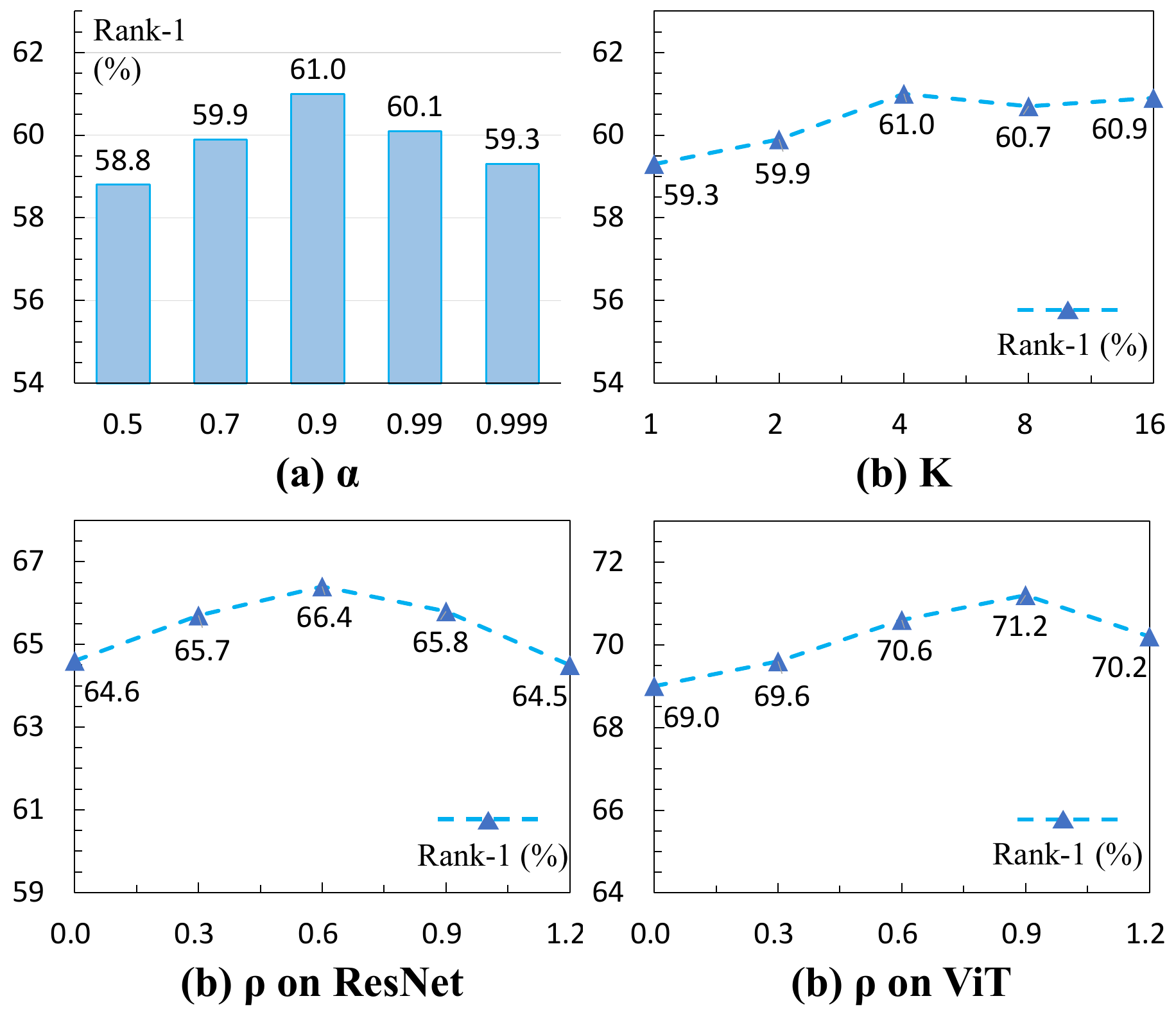}
    \caption{Effect of the hyper-parameter $\alpha$, $K$ and $\rho$ on PRCC in the cloth-changing setting. We first evaluated the $\alpha$ and $K$ without the Disentangle Regularization, and then evaluated the $\rho$ under $\alpha$ and $K$ are set to 0.9 and 4.
    }
    \hfill
    \label{fig:param}
    \vspace{-3mm}
\end{figure}

As shown in Table~\ref{tab:label_robust}, compared with no causal intervention, all settings achieve substantial improvements. Even under 100\% within-ID clothing-label randomization, the performance only drops by 1.8 percentage points in Rank-1 compared to using ground-truth clothing labels.
This indicates that, even when labels are fully degraded, intervention remains effective as long as label operations are restricted within each identity. Based on this, we use a simpler fully clothing-label-free alternative: using identity labels as grouping units, which achieves performance close to the fully randomized setting.
We further apply DBSCAN~\cite{ester1996density} clustering within each identity to recover finer-grained clothing groups, achieving 60.5\% Rank-1 and 59.4\% mAP, which is close to the ground-truth performance.
These results demonstrate that our causal intervention framework has weak dependence on clothing-label quality.

Note that the Disentangle Regularization is designed to extract clothing representations and is more sensitive to clothing-label quality. When clothing labels are unavailable, a pre-trained human parsing model can be used to assist clothing feature extraction, achieving comparable performance.

\subsection{Hyper-parameters Analysis}
We analyzed the influences of some key hyper-parameters on the PRCC dataset cloth-changing setting, including $\alpha$, $K$ and $\rho$. 
Specifically, $\alpha$ denotes the memory coefficient of Equation~\ref {eq:alpha} to control the update speed of the Confounder Dictionary, $K$ denotes the sample times of Equation~\ref {eq:f2} to control the richness of confounder during the intervention process.
We first evaluated the $\alpha$ from \{0.5, 0.7, 0.9, 0.99, 0.999\} and $K$ from \{1, 2, 4, 8, 16\} without the Disentangle Regularization to avoid the influence of $\rho$. 
As shown in Figure~\ref{fig:param} (a), setting $\alpha$ = 0.9 yields the highest performance. It strikes a balance between update speed and stability.
As shown in Figure~\ref{fig:param} (b), setting $K$ = 4 is sufficient, and a larger $K$ is unnecessary.
The $\rho$ is the margin parameter of our feature separation loss $\mathcal{L}_{fs}$ in Equation~\ref {eq:fo}, which will affect the effectiveness of the Disentangle Regularization.
We evaluated the $\rho$ under $\alpha$ is set to 0.9 and $K$ is set to 4.  
Due to the differences in the feature spaces of CNNs and Transformers, they require different margin parameters.
As shown in Figure~\ref{fig:param} (b) and (c), the performance rises and then decreases as the $\rho$ increases, and the performance reaches a peak when $\rho$ is set to 0.6 for the ResNet backbone and set to 0.9 for the ViT backbone.
We did not search for the best hyper-parameters for each dataset, but instead utilized these settings for other CC-ReID datasets.

\begin{figure*}[!t]
    \centering
    \includegraphics[width=0.80\linewidth]{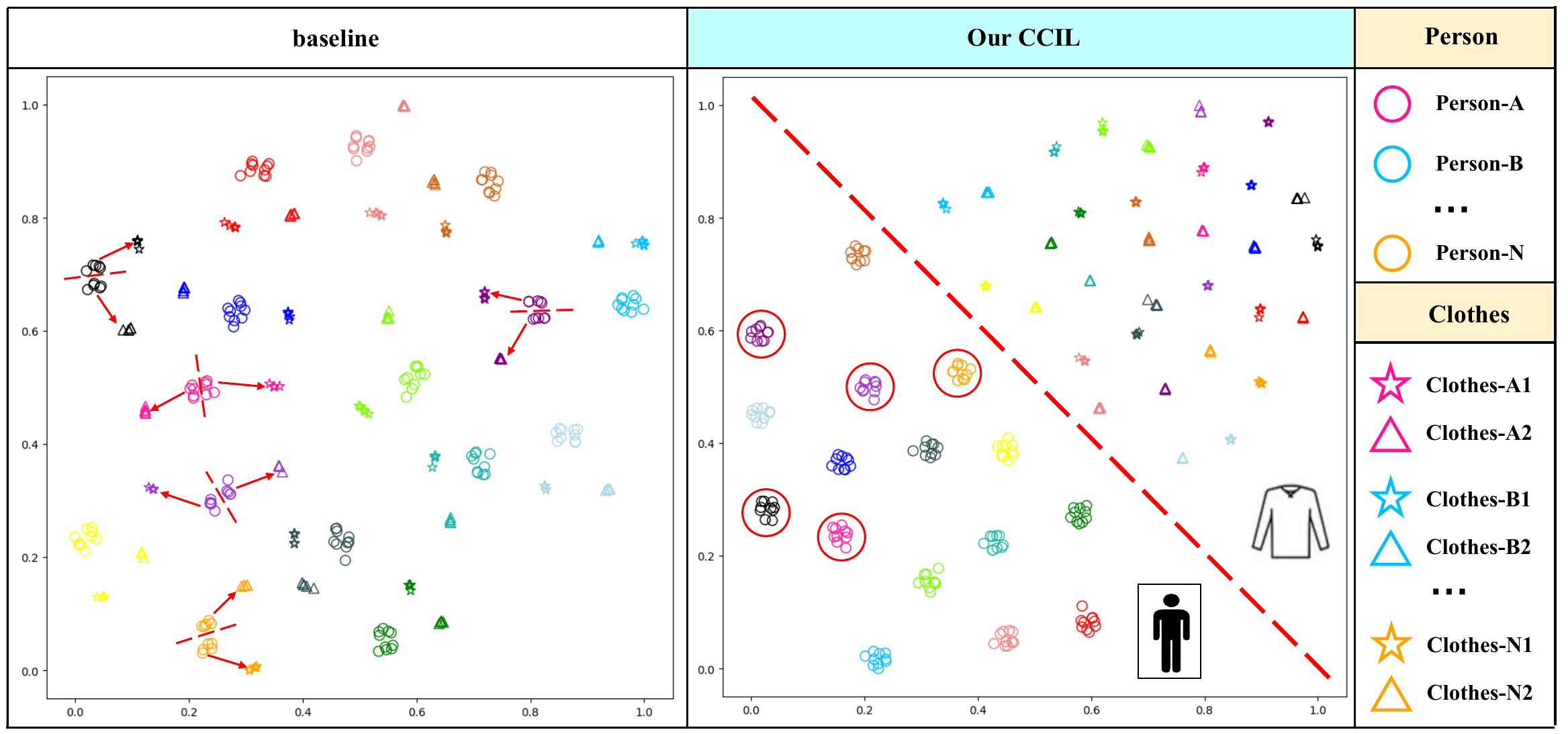}
    \caption{t-SNE~\cite{van2008visualizing} visualization of the distributions of image features $f_{img}$ and clothes features $f_{clt}$ on the PRCC dataset. 
    Different colors represent different identities. The circle represents a person, and the triangle and star represent the person's first and second sets of clothes, respectively.
    }
    \hfill
    \label{fig:tsne}
    \vspace{-3mm}
\end{figure*}

\begin{figure}[!t]
    \centering
    \includegraphics[width=0.85\linewidth]{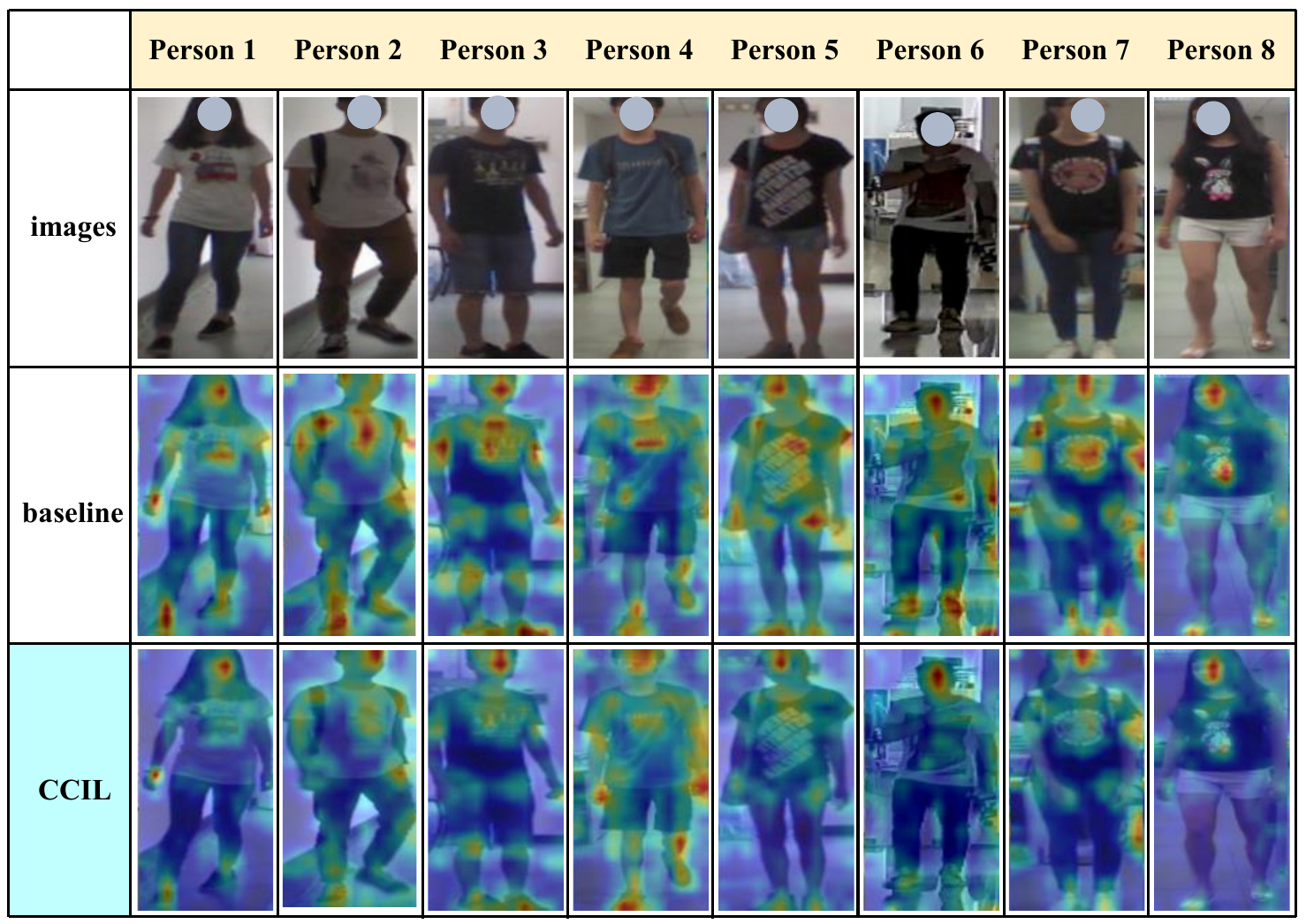}
    \caption{Visualization of the activation feature maps on PRCC, and LTCC datasets. In each triplet of a column, the first line indicates the original image, the second line corresponds to the baseline model, and the third line corresponds to our CCIL model.
    }
    \hfill
    \label{fig:cam}
    \vspace{-3mm}
\end{figure}

\begin{figure}[!t]
    \centering
    \includegraphics[width=0.75\linewidth]{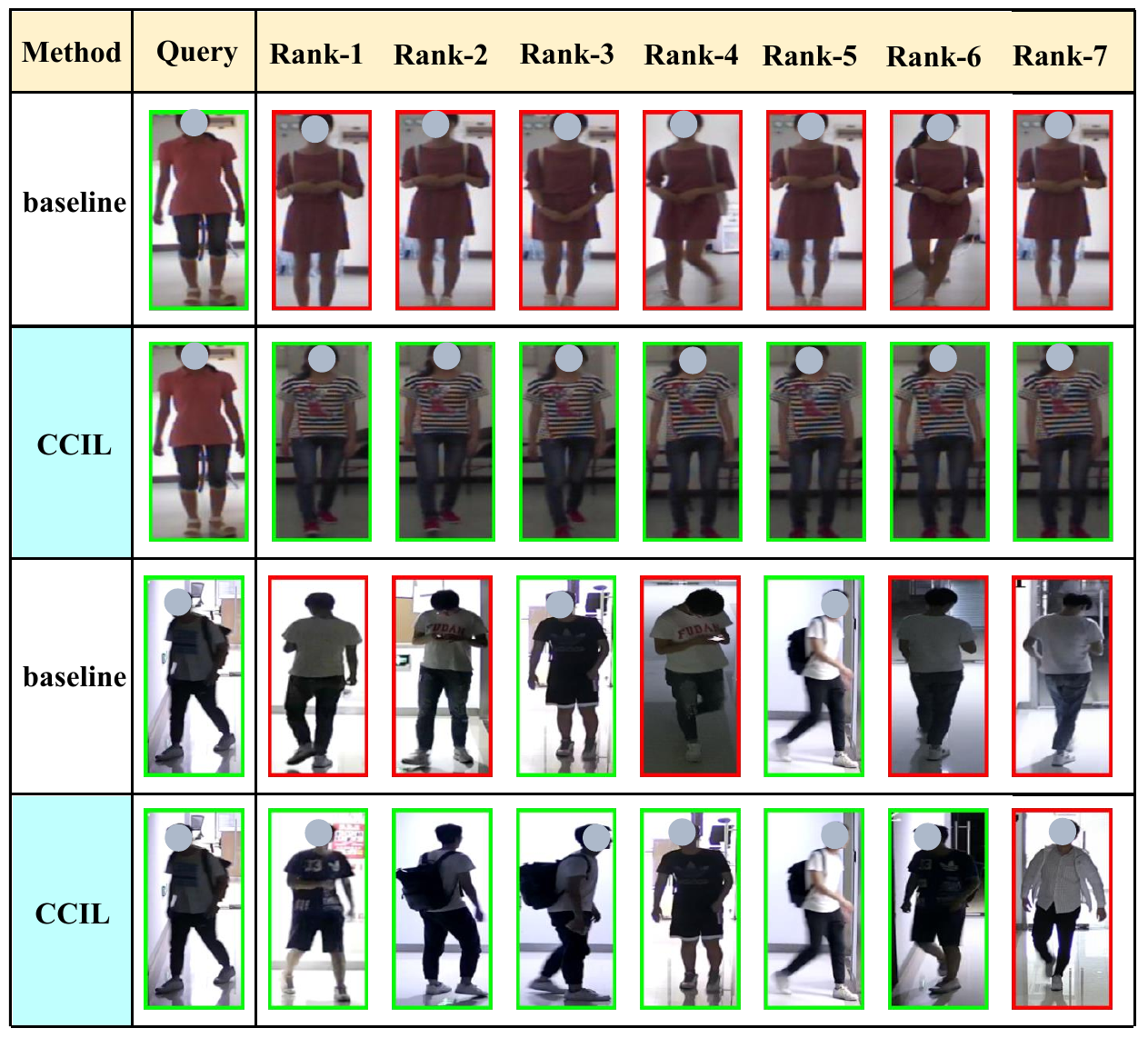}
    \caption{Visualization of the retrieval ranking lists on PRCC and LTCC datasets. The green boxes represent correct retrieval results, and the red boxes represent incorrect retrieval results. 
    }
    \hfill
    \label{fig:rank}
    \vspace{-4mm}
\end{figure}

\section{Visualization}

\subsection{Visualization of the feature distribution.}

Figure~\ref{fig:tsne} shows t-SNE~\cite{van2008visualizing} visualization results for image features $f_{img}$ and clothing features $f_{clt}$ on the PRCC dataset.

In the feature distribution of the baseline model, image features (circles) and clothing features (triangles and stars) of the same identity (represented by color) are mixed together.
It shows that the baseline extracted $f_{img}$ and $f_{clt}$ have a strong correlation.
On the contrary, our method perfectly distinguishes person features and clothing features, and it is easy to find a line to separate them in the figure.

In addition, some clusters of the baseline model are influenced by clothing, resulting in their division into two parts, whereas our method exhibits a more compact intra-class feature distribution.

\subsection{Visualization of the activation feature maps.}
As shown in Figure~\ref{fig:cam}, we visualize activation feature maps of several images on PRCC and LTCC datasets.
In each triplet of a column, the first line indicates the original image.
The second and third lines present the activation maps of the baseline model and our CCIL model, respectively.
The activation maps indicate that our method focuses on ID-related cues, such as the head, joints, hands, and feet, corresponding to stable causal relationships in recognition, which remain consistent regardless of variations in clothing or scenes.

Our method, in comparison to the baseline model, neglects clothing-related regions, particularly prominent patterns on clothing.
Meanwhile, our method still allocates slight attention to certain areas of clothing, as these regions also contain effective cues, such as body shape and joints. 
Completely ignoring the clothing areas may result in the loss of certain identity cues, leading to overcorrection.
The visualization of activation feature maps reflect the superiority of our method in learning clothing-invariant features.

\subsection{Visualization of the retrieval results.}
Figure~\ref{fig:rank} illustrates some retrieval results of our proposed CCIL (2-$nd$ and 4-$th$ lines) versus baseline (1-$st$ and 3-$rd$ lines) on PRCC and LTCC datasets.
The green boxes are the positive search results and the red boxes mean negative results.
The retrieval results show that our method can retrieve some harder positive targets that the baseline model cannot, even if the style, color, and texture of their clothes are quite different from the query.
 
\section{Conclusion}
This paper studies the cloth-changing person Re-identification (CC-ReID) task from a novel causal perspective.
We argue the spurious correlation between clothing and identity in the dataset may interfere with the crucial clothes-invariant feature learning for CC-ReID.
To address this issue, we propose a Causal Clothes-Invariant Learning (CCIL) by modeling the intervention probability $P(Y|do(X))$.
The proposed method includes three modules that are complementary to each other and train the model under the causal intervention framework, achieving better clothes-invariant features.
The extensive experiments and visualizations on multiple CC-ReID datasets validated the effectiveness and advantage of our
method.

\bibliographystyle{IEEEtran}
\bibliography{main}

\end{document}